\documentclass[10pt,twocolumn,letterpaper]{article}
\usepackage[accsupp]{axessibility}
\usepackage{iccv}
\usepackage{times}
\usepackage{epsfig}
\usepackage{graphicx}
\usepackage{amsmath}
\usepackage{amssymb}
\usepackage{booktabs}
\usepackage{multirow}
\usepackage{color}
\usepackage{bbm}
\usepackage{tabu}
\usepackage{threeparttable}
\usepackage[table]{xcolor}

\usepackage[pagebackref=true,breaklinks=true,letterpaper=true,colorlinks,bookmarks=false]{hyperref}

\iccvfinalcopy % *** Uncomment this line for the final submission

\def\iccvPaperID{****} % *** Enter the ICCV Paper ID here

\ificcvfinal\fi

\begin{document}

%%%%%%%%% TITLE
\title{Prototypical Kernel Learning and Open-set Foreground Perception for Generalized Few-shot Semantic Segmentation}

% \author{First Author\\
% Institution1\\
% Institution1 address\\
% {\tt\small firstauthor@i1.org}
% % For a paper whose authors are all at the same institution,
% % omit the following lines up until the closing ``}''.
% % Additional authors and addresses can be added with ``\and'',
% % just like the second author.
% % To save space, use either the email address or home page, not both
% \and
% Second Author\\
% Institution2\\
% First line of institution2 address\\
% {\tt\small secondauthor@i2.org}
% }
\author{Kai Huang, Feigege Wang, Ye Xi, Yutao Gao\\
Alibaba Group\\
{\tt\small zhouwan.hk, feigege.wfgg, yx150449, yutao.gao@alibaba-inc.com}
}

\maketitle
% Remove page # from the first page of camera-ready.
\ificcvfinal\thispagestyle{empty}\fi

%%%%%%%%% ABSTRACT
\begin{abstract}
   Generalized Few-shot Semantic Segmentation (GFSS) extends Few-shot Semantic Segmentation (FSS) to simultaneously segment unseen classes and seen classes during evaluation. Previous works leverage additional branch or prototypical aggregation to eliminate the constrained setting of FSS. However, representation division and embedding prejudice, which heavily results in poor performance of GFSS, have not been synthetical considered. We address the aforementioned problems by jointing the prototypical kernel learning and open-set foreground perception. Specifically, a group of learnable kernels is proposed to perform segmentation with each kernel in charge of a stuff class. Then, we explore to merge the prototypical learning to the update of base-class kernels, which is consistent with the prototype knowledge aggregation of few-shot novel classes. In addition, a foreground contextual perception module cooperating with conditional bias based inference is adopted to perform class-agnostic as well as open-set foreground detection, thus to mitigate the embedding prejudice and prevent novel targets from being misclassified as background. Moreover, we also adjust our method to the Class Incremental Few-shot Semantic Segmentation (CIFSS) which takes the knowledge of novel classes in a incremental stream. Extensive experiments on PASCAL-5$^i$ and COCO-20$^i$ datasets demonstrate that our method performs better than previous state-of-the-art.
\end{abstract}

%%%%%%%%% BODY TEXT
\section{Introduction}
\label{sec:intro}
Few-shot semantic segmentation \cite{FirstFSS_BMVB_2017, CANet_CVPR_2019, boudiaf_CVPR_2021, lang2022learning} aims to handle the situation of only sparse annotated data available, which perplexes most normal semantic segmentation methods. Inherited from the superiority of few-shot learning, the FSS methods can quickly be expended to unseen domains with only few labeled samples. Typically, most FSS methods are built on episode-based meta-learning \cite{snell_2017_NIPS}, each episode data contains a support set and a query set shared with same classes. In general, the support set consists of few support images with pixel-wise annotations. The models are supported to learn a generic meta-learner with abundant episodic data sampled from the base classes, and adopt to novel classes with the guidance of the support set. However, the majority of current FSS methods can not escape from the limitation of support set, in which specific as well as single novel class annotations are needed corresponding to query images during the inference. Another weakness is that models can only segment the specific novel class of support set while lack the identification of base classes or multi-novel classes. The recent work \cite{lang2022learning} leverages the prediction of base classes with a base learner, and thus gives a mask region, which instructs the area does not need to be segmented, to guide the recognition of novel concepts. Despite the significant improvement has been made in base classes, the issue of heavily relying on the support samples with the same classes for each query image still exists.

\begin{figure}[t!]
    \centering
    \setlength{\abovecaptionskip}{0cm}
    \setlength{\belowcaptionskip}{-0.5cm}
	\includegraphics[width=0.5\textwidth]{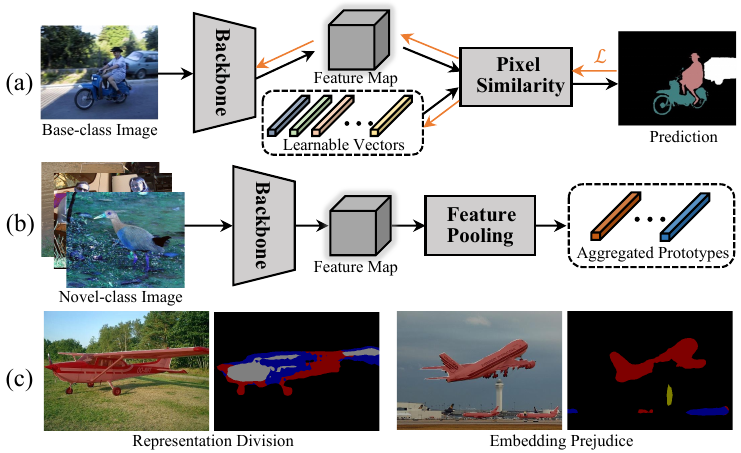}
	\caption{Illustration of our motivation. (a) shows the generation of base-class representation vectors, it learns from abundant base-class images. (b) is the process of novel-class prototypes, which are aggregated by the lightweight feature pooling since few samples available. (c) demonstrates the representation division and embedding prejudice problem existed in current GFSS.}
 	\label{fig1_inference}
\end{figure}
To this end, Generalized Few-shot Semantic Segmentation (GFSS) was proposed \cite{tian2022generalized} where novel classes are aforehand registered to the vanilla model learned from the base classes, making it possible to segment the base and novel simultaneously without extra information in the inference phase. The typical similarity metric enables a more flexible way to expand novel classes with few annotated samples above the base classes. It can directly perform the inference the same as normal segmentation models, with no need of the prerequisite which informs the target classes contained in query images. Regardless, there are still two challenges in GFSS as shown in Fig.\ref{fig1_inference}(c): 1) The representation division between base-class and novel-class blocks the accurate predictions, where the former are learned from sufficient data and the latter are aggregated from limited samples with feature pooling approaches; 2) the prejudice of feature embedding makes the novel-class pixels have the tendency to be identified as varied background, since the feature extractor is trained in base classes. 

In this work, we propose to joint prototypical kernel learning and open-set foreground perception to address the above-mentioned problems. It exploits a base-class kernel update schema by applying the pixel assembled feature of each input image to extra supervise the update of base-class kernels. Apart from the normal update with abundant base-class samples, those kernels are endowed with the ability of prototypical representation by the prototypical kernel update, which is consistent with the knowledge aggregation of novel-class with few labeled data. To mitigate the feature embedding preconception, we leverage the common feature patterns cross multi-class targets by an open-set foreground perception, which serves the purpose to offer the class-agnostic as well as open-set foreground prediction. Finally, a conditional bias based inference is proposed to assemble the class-wise prediction and the binary foreground prediction. Furthermore, we expand our method from the GFSS to class incremental few-shot segmentation which registers novel classes with an incremental stream.

In summary, our primary contributions can be summarized as: 1) we present a prototypical kernel learning for maintaining the representation consistency between base-class and novel-class by the notion of kernel; 2) we propose to learn the class-agnostic foreground perception cross the multi-class targets to mitigate the feature embedding preconception, and devise a conditional bias based inference for ensemble prediction; 3) extensive experiments on GFSS and CIFSS demonstrate that the proposed method achieves higher performance compared with other approaches.

%-------------------------------------------------------------------------
\section{Related Work}
\label{sec:relatedwork}
{\bf Few-shot Segmentation} aims to give a dense pixel prediction for the query images with only a few annotated support images available. The most recent methods adopt two-branch paradigm, \emph{i.e.}, a support branch for target classes prototypical representation and a query branch for feature pairs comparison. Typically, the support branch is performed by the mask-involved feature pooling, such as masked average pooling \cite{Siam_ICCV_2019}, dynamic kernel generation \cite{liu2022dynamic}. The query branch tries to strengthen the metric ability between support prototypes and query feature, with directly pixel similarity \cite{Dong_BMVC_2018}, convolutional correlation \cite{liu2022dynamic, HSNet_ICCV_2021}, attention-based interaction \cite{wu_ICCV_2021, zhang2021few} and so on. However, limited by the instance-based feature comparison of query branch, those methods can not handle on the GFSS scene, in which both base classes and novel classes are included to validation. BAM \cite{lang2022learning} introduces a base learner module for highly reliable base classes segmentation, and further incorporates into the meta prediction with an ensemble module. Nevertheless, it still needs specific support data and only works with single novel class for each query task. CAPL \cite{tian2022generalized} proposes a novel class registration phase to register the novel information to base classifier, making it possible to be qualified for the segmentation of both base and multi-novel classes without additional support samples.

{\bf Prototypical Learning} is meant to excavate or utilize the representative examples directly with the exemplar-driven pattern. Owing to the needs of representation with limited data, the prototypical learning methods have been widely used in many few-shot tasks \cite{snell_2017_NIPS,huang2021pseudo,Dong_BMVC_2018,zhang2021prototypical}, especially few-shot segmentation \cite{PANet_ICCV_2019, zhou2022rethinking}. A typical delegate is the prototype-based framework, in which annotated support data is class-wise aggregated as prototypes and then guide the segmentation of query samples. The prototypical learning of FSS can be roughly divided into implicit and explicit. The implicit methods mainly focus on the embedding learning and tend to use the feature pooling approaches, such as masked average pooling \cite{Siam_ICCV_2019}, attention-based aggregation \cite{wu_ICCV_2021} and kernel generation \cite{liu2022dynamic}. While the explicit way is to construct learnable prototypes, so that the labeled instances \cite{zhou2022rethinking} or meta episode data \cite{yang_ICCV_2021} can react it directly.

{\bf Foreground Modeling} has gained rapid development with the effectiveness of targets detection. The most representative methods are class activation mapping and its variants \cite{zhou2016learning, selvaraju2017grad, wang2020score}, which are early used for feature visualization \cite{zhou2016learning} and further as the rough foreground mask to initialize the progressive inception of many weak-supervised approaches \cite{du2022weakly, chen2022self}. Later, salient detection \cite{qin2019basnet, zhao2019egnet, wang2019salient} has been raised for binary segmentation tasks, with the strong ability of flexible saliency perception. The mainstream salient works only consider the intra interaction on each single image separately, which suffer from obscure and coarse results. Co-salient \cite{zhang2019co, fan2021group} extends the separate mode to co-exploration within the image group, while extra information, \emph{e.g.}, edge \cite{fan2021re}, class-wise gradient \cite{zhang2020gradient}, is always needed for discriminative feature learning. In this work, we propose to learn the common foreground patterns cross multi-class group instances, so as to directly promote to the detection of novel targets without class limitation.

\begin{figure*}[ht!]
    \centering
    \setlength{\abovecaptionskip}{0cm}
    \setlength{\belowcaptionskip}{-0.2cm}
	\includegraphics[width=\textwidth]{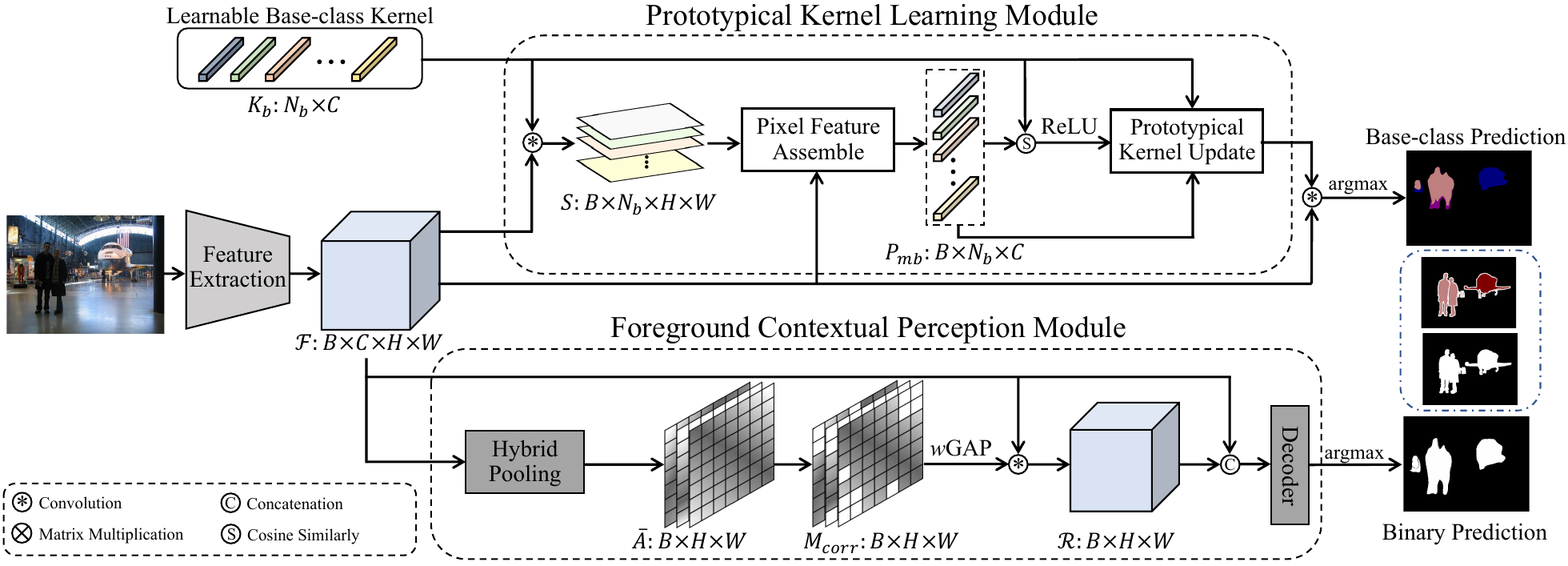}
	\caption{Overview of training framework. For the upper branch, the learnable base-class kernels are optimized by the Prototypical Kernel Learning (PKL). The down branch, named as Foreground Contextual Perception (FCP), is constructed to learn the open-set foreground detection by leveraging the contextual feature patterns cross multi-class targets. Batch is omitted in PKL for better display.}
 	\label{fig: overview_model}
\end{figure*}

\section{Method}
\label{sec:method}
\subsection{Task Formulation}
Generalized few-shot segmentation is designed to quickly integrate unseen classes into a segmentation model with only a few annotated data. It aims to actualize segmentation tasks on novel classes with limited labeled samples and base classes with sufficient labeled data simultaneously. GFSS model learns the basal knowledge from the base classes set $C_{\emph{base}}$, and novel classes set $C_{\emph{novel}}$ is registered to this base model with a few labeled data. There are no overlap between the base and novel classes set, \emph{i.e.}, $C_{\emph{base}} \cap C_{\emph{novel}}=\emptyset$. In the inference process, the model should be qualified to the segmentation of the union set $C_{\emph{base}} \cup C_{\emph{novel}}$. Different from the normal FSS which are specific to the modeling of few-shot learner, GFSS is proposed to explore the possibility of empowering the few-shot learning ability to a vanilla segmentation model.

\subsection{Overview}
The overall architectures for training and inference are shown in Fig. \ref{fig: overview_model} and Fig. \ref{fig4}, which mainly consists of three components: Prototypical Kernel Learning (PKL), Foreground Contextual Perception (FCP) and Conditional Bias Based Inference (CBBI). The feature of input is respectively fed to two branches. For the upper branch, we employ the PKL to dynamically get the refined base-class kernels specific to the assembled feature of input image, and the class-wise predictions are produced by applying those kernels to the pixel features. For the down branch, the binary foreground prediction is decoded by the FCP which incorporates relevant contextual patterns cross multi-classes. The final prediction masks are obtained by combining the predictions of these two branch with the CBBI.

\subsection{Prototypical Kernel Learning Module}
Inspired by the dynamic kernel \cite{gao2019deformable, wang2020solov2, zhang2021k} in semantic segmentation, which possess better representation ability than convolution kernels with flexible structures \cite{wu2018dynamic} or roles \cite{li2022video}, we extend this to GFSS task by adopting prototypical kernel learning to maintain the representation consistency of base-class and novel-class. 

Speciﬁcally, $N_{b}$ learnable kernels $K_{b} \in \mathbb R^{N_{b}\times C}$are constructed to represent the knowledge of $N_{b}$ base classes, in which the convolution between these kernels and feature maps is taken on the segmentation basis. Concretely, given the input feature maps $\mathcal{F}\in \mathbb R^{B\times C \times H \times W}$ of $B$ input images, the semantic segmentation output is formulated as:
\begin{equation}\label{Eq.1}
    S=\text{softmax}(K_{b}*\mathcal{F}), S\in \mathbb R^{B\times N_{b} \times H \times W},
\end{equation}
where $C$, $H$, and $W$ indicate the number of channels, height, and width of the feature map respectively, and $N_{b}$ is the number of classes in base-class set. 

\noindent \textbf{Pixel Feature Assembling.} As the segmentation output of each kernel can reflect the class-wise pixel-to-prototype mapping, the pixel feature assembling is performed by aggregating the input feature with the pseudo semantic segmentation masks as:
\begin{equation}
\setlength{\abovedisplayskip}{7pt}
\setlength{\belowdisplayskip}{7pt}
P_{\emph{mb}}=\sum_{x}^{H} \sum_{y}^{W} S(x, y) \cdot \mathcal{F}(x, y), P_{\emph{mb}}\in \mathbb R^{B\times N_{b}\times C}.
\end{equation}
The generation of $P_{\emph{mb}}$ are analogous to the novel-class feature aggregation with few labeled samples, which takes the ability of prototypical representation.
% which is illustrated in Fig. \ref{fig1_inference}(b) and 

\noindent \textbf{Prototypical Kernel Update.} Ordinarily, the kernel of base-class learn the general perception from abundant annotated samples, it would be broader and more inclusive than the novel-class kernels learned from few data, which exists representation division. Those ambiguous instances, \emph{e.g.}, the novel classes samples, are more predisposed to be caught by the base-class kernels. Thus, the prototypical kernel update is proposed to adapt the base-class kernels to few-shot task with extra one-off update specific to each input:
\begin{equation}\label{Eq.3}
    \tilde{K}_{b}=K_{b}-\alpha \nabla_{K_{b}} \mathcal{L}\left(K_{b}\right)
\end{equation}
where $\nabla_{K_{b}} \mathcal{L}\left(K_{b}\right)=\nabla_{K_{b}} (K_b-P_{\emph{mb}}^i)^\text{2}$ is the gradient descent of the mean squared error loss between $K_{b}$ and the assembled feature of \emph{i}-th images $P_{\emph{mb}}^i$, $\alpha$ is the adaptation learning rate, which denotes as kernel-to-prototype similarity with the restraint of ReLU function that insures constant positive:
\begin{equation}
    \alpha=\text{ReLU}(\text{cosine}(K_b, P_{mb}^{i})), \alpha \in \mathbb{R}^{1 \times N_b}.
\end{equation}
The adaptation learning rate is designed to adaptive filter out ineffective gradient signals caused by those classes do not exist in this image. Since the classes existed in input image are a subset of base-class set, the pixel feature assembling of those missing classes is invalid and will lead to incorrect kernel update. The adaptation learning rate will decay to near zero to block the ineffective update as the kernel-to-prototype similarities are low. Finally, The predicted segmentation mask is obtained by: 
\begin{equation}
\setlength{\abovedisplayskip}{7pt}
\setlength{\belowdisplayskip}{7pt}
M(x, y)=\underset{i}{\arg \max } \tilde{S}^{i}(x, y),
\end{equation}
where $\tilde{S}^{i}(x, y)$ is the score map between the updated base-class kernels and the feature maps followed by Eq. (\ref{Eq.1}).

\subsection{Foreground Contextual Perception Module}
%specific input
The aforementioned PKL module aims to adjust the general base-class kernels to few-shot scenario with assembled prototypical feature, so that the representation divergence between base and novel classes is mitigated. While the perception ability of novel foreground targets remains poor with the limitation of backbone training, where only the base-class data is sufficiently accessible. Inspired by the episode-based paradigm in various few-shot learning \cite{vinyals2016matching, FirstFSS_BMVB_2017, huang2021pseudo}, which can boost the model to adapt to unknown scenes, we propose a foreground contextual perception module to perform the open-set foreground perception. The batch group of images sampled from the base categories set can be considered as a pseudo episode, in which each batch group is randomly mixed with the samples of different base-class targets. The FCP module aims to mine common foreground patterns cross multi-class samples with abundant pseudo episode, thus possesses the ability of class-agnostic perception and prevents novel targets from being misclassified as background. 

\begin{figure}[t!]
    \centering
    \setlength{\abovecaptionskip}{0cm}
    \setlength{\belowcaptionskip}{-1.2cm}
	\includegraphics[width=0.5\textwidth]{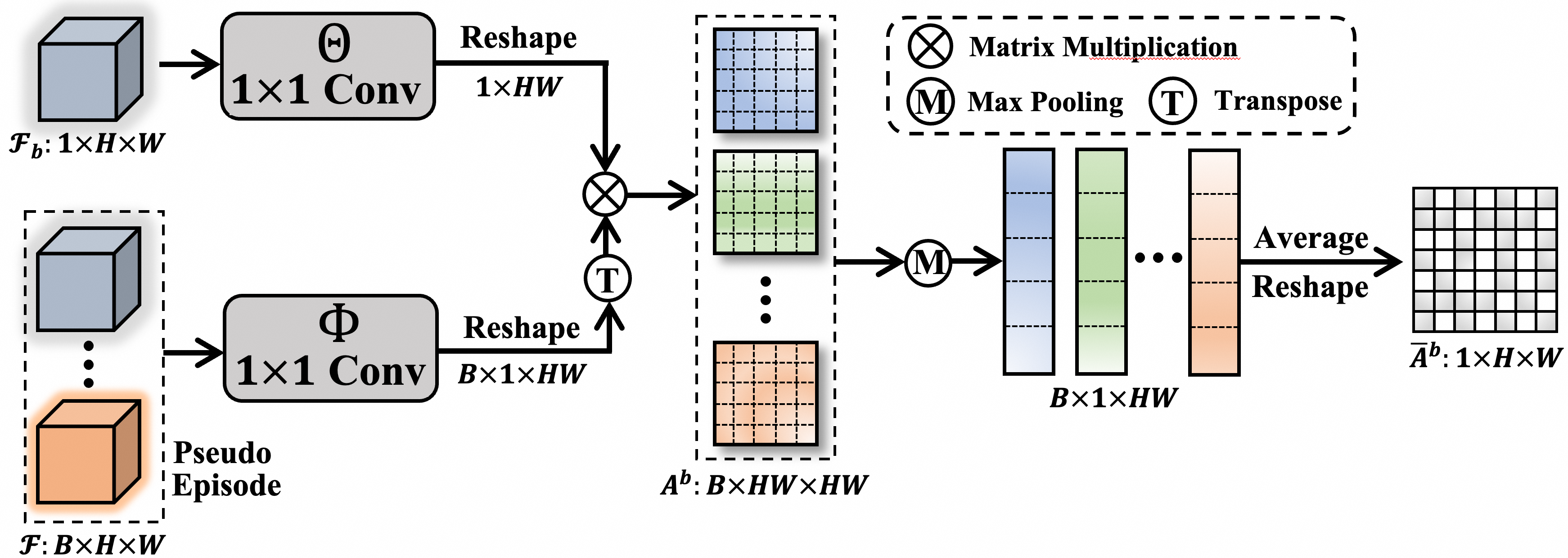}
	\caption{Illustration of the Hybrid Pooling in FCP module.}
 	\label{fig3}
\end{figure}

\noindent \textbf{Hybrid Pooling.} As shown in Fig.\ref{fig3}, the feature maps of pseudo episode are first passed by two independent linear mappings, and the pixel-wise correlation is calculated within the feature map groups:
\begin{equation}\label{Eq.6}
\setlength{\abovedisplayskip}{7pt}
\setlength{\belowdisplayskip}{7pt}
   A^{b} = \Phi(\mathcal{F})^{T}\Theta(\mathcal{F}_{b}) \in \mathbb R^{B \times HW\times HW},
    %\label{eq: linear_projection}
\end{equation}
where $\Phi$ and $\Theta$ represent $1\times 1$ convolution layers, $\mathcal{F}_{b}$ is the \emph{b}-th feature map, and $A^{b}$ denotes the pixel-wise correlation between the \emph{b}-th feature map and whole pseudo episode. The first dimension of $A^{b}$ indicates all images of pseudo episode, while the second and third dimensions mean the total feature pixels of each image in pseudo episode and the total feature pixels of query. The reshaping and transposing operations of matrix dimension are omitted for the simplification of presentation. A max pooling and average pooling are separately acted on the second and first dimensions to obtain a more representative correlation $\bar{A} \in \mathbb R^{B\times H\times W}$. It is worth mentioning that performing max pooling on pixel dimension can help to find out discriminative correlation pairs, and average pooling on image dimension is supported to learn the common foreground feature patterns. 

\noindent \textbf{Foreground Correlation Responses.} The correlation normalized by a softmax layer is truncated as the correlation mask $M_{\emph{corr}}$:
\begin{equation}\label{Eq7}
\setlength{\abovedisplayskip}{7pt}
\setlength{\belowdisplayskip}{7pt}
   M_{\emph{corr}}(h,w) = \mathbbm{1}(\bar{A}(h,w) > \text{0.5}),
\end{equation}
where $\mathbbm{1}(\ast)$ is the binary indicator that outputs 1 when $\ast$ is true 0 otherwise, $h$ and $w$ are the height index and width index with the limitation of $H$ and $W$. Although the correlation mask can embody the most discriminative feature pixel position, the relation response of other positions are discarded with the consideration of computational complexity. A sharp correlation information may carries more hard noisy, which blocks the learning of FCP module. Therefore, the spacial weighted Global Average Pooling (\emph{w}GAP) \cite{CANet_CVPR_2019, Zhang_CVPR_2021} of the correlation mask together with the normalized original feature maps is introduced to obtain a more robust correlation prototypes as:
\begin{equation}\label{Eq8}
\setlength{\abovedisplayskip}{7pt}
\setlength{\belowdisplayskip}{7pt}
    p_{\emph{wGPA}} = \frac{\sum\nolimits_{h,w}\lVert\mathcal{F}(h,w)\rVert_{2} M_{\emph{corr}}(h,w)}{\sum\nolimits_{h,w}M_{\emph{corr}}(h,w)} \in \mathbb R^{B\times \emph{C}},
\end{equation}
where $\lVert \ast \rVert_{2}$ is the L2 normalization. The correlation prototypes reflect the positive relations between the pixel and whole image pixels within the pseudo episode. The refined correlation responses can be obtained by applying the correlation prototypes to the original feature maps as:
\begin{equation}
\setlength{\abovedisplayskip}{7pt}
\setlength{\belowdisplayskip}{7pt}
   \mathcal{R}(h,w) = \frac{1}{B}\sum_{i=1}^Bp_{\emph{wGPA}}\lVert\mathcal{F}(h,w)\rVert_{2}.
    %\label{eq: linear_projection}
\end{equation}
Compared to the pixel-wise correlation referred to Eq. (\ref{Eq.6}), the refined correlation responses are more robust with fewer sharp noises, as well as more accurate foreground perception. The final binary foreground masks are predicted by decoding the residual features:
\begin{equation}
\setlength{\abovedisplayskip}{7pt}
\setlength{\belowdisplayskip}{7pt}
  M_{f} = \mathbb{D}(\mathcal{F} + \mathcal{F} \times \mathcal{R}),
\end{equation}
where $\mathbb{D}$ represents a correlation decoder built with the four cascaded convolutional layers.

\subsection{Training Loss and Novel Classes Registration}
\label{se.3.5.}
\noindent \textbf{Training Loss.} The loss for training our model consists of two parts corresponding to the PKL module and FCP module respectively. The loss function for PKL module is illustrated as:
\begin{equation}
\setlength{\abovedisplayskip}{7pt}
\setlength{\belowdisplayskip}{7pt}
\mathcal{L}_{\emph{ce}}=\frac{1}{H W} \sum_{x, y} \sum_{p^i \in P_{ts}} \mathbbm{1}\left[M^{x, y}=i\right] \log \hat{S}_{c}^{x, y},
\end{equation}
where $\mathbbm{1}(\ast)$ is the binary indicator same as Eq. (\ref{Eq7}). We adopt the IoU loss \cite{qin2019basnet} to train the FCP module as:
\begin{equation}
\setlength{\abovedisplayskip}{7pt}
\setlength{\belowdisplayskip}{7pt}
\mathcal{L}_{\emph{iou}}= \frac{1}{B}\sum_{b=1}^B(1-\frac{\hat{Y} \cap Y}{\hat{Y} \cup Y}),
\end{equation}
where $\hat{Y}$ is the binary predictions and $Y$ is the ground truth. Finally, the total end-to-end loss over the batch is:
\begin{equation}
\setlength{\abovedisplayskip}{7pt}
\setlength{\belowdisplayskip}{7pt}
\mathcal{L}_{\emph{total}} = \lambda \sum_{b \in B} \mathcal{L}_{\emph{ce}}^{b} + (1-\lambda) \mathcal{L}_{\emph{iou}},
\end{equation}
where $\lambda$ is utilized to balance the loss functions. 

\noindent \textbf{Novel Classes Registration.} After training, the novel-class kernels $K_{\emph{n}} \in \mathbb R^{N_{\emph{n}} \times C}$ can be obtained by passing the support images of novel classes to the frozen feature extraction, following with the class-wise averaged \emph{w}GAP referred to Eq. (\ref{Eq8}). Then, the novel classes knowledge is registered to the model by concatenating the novel-class kernels to the updated base-class kernels:
\begin{equation}
\setlength{\abovedisplayskip}{7pt}
\setlength{\belowdisplayskip}{7pt}
    K_{\emph{w}} = \emph{Concat}([K_{\emph{b}}, K_{\emph{n}}]), K_{\emph{w}} \in \mathbb R^{(N_b+N_n)\times C}.
\end{equation}

\begin{figure}[t]
    \centering
    \setlength{\abovecaptionskip}{0cm}
    \setlength{\belowcaptionskip}{-0.5cm}
	\includegraphics[width=0.5\textwidth]{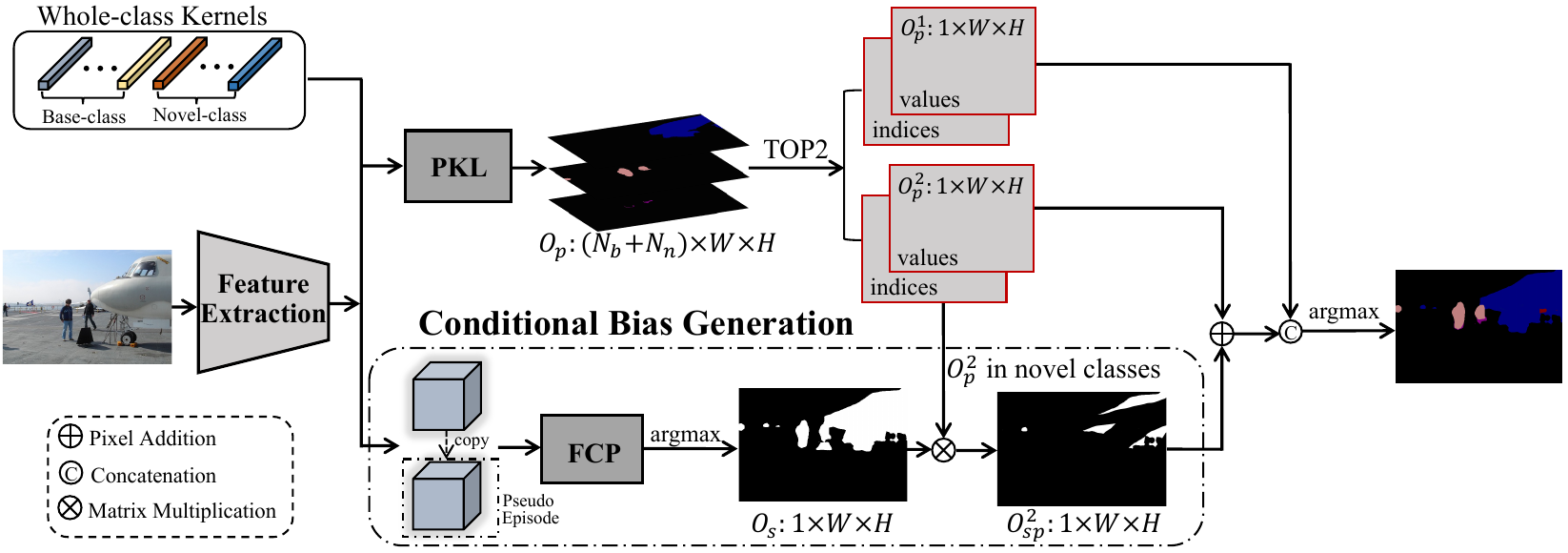}
	\caption{Framework of the CBBI. We perform the FAH module with only itself contained in the pseudo episode.}
 	\label{fig4}
\end{figure}

\subsection{Conditional Bias Based Inference}
Given the input test image $I$, the final decision of position $\mathbf{p}$ is defined as follows:
\begin{equation}
\setlength{\abovedisplayskip}{7pt}
\setlength{\belowdisplayskip}{7pt}
    O_{I}(\mathbf{p}) = \begin{cases} \text{max}(c_{\text{1st}},c_{\text{2nd}}+\emph{b}), & O_s(\mathbf{p}) == 1 \\ c_{\text{1st}}, & \text {otherwise} \end{cases},
\end{equation}
where $c_{\text{1st}}$ and $c_{\text{2nd}}$ are the first and second candidates according to the class-wise similarity, $O_s$ indicates the output of FCP module and $b$ is the constant of bias.

As shown in Fig. \ref{fig4}, the refined meta specific prototypes are obtained by feeding the feature map of test image together with the class-wise prototypes to the PKL module. With the concerns of consistency and noise accumulation, we only update the base-class prototypes. Then, the class-wise predictions $O_p$ are calculated by the cosine similarity between the updated whole-class prototypes and query feature maps. The first and second candidate masks are the value of Top2 maps along with the class dimension of $O_p$. 

For the branch of conditional bias generation, the open-set foreground mask $O_s$ can be obtained by the FCP module. A pointwise multiplication between it and the second candidate mask is adopted to obtain the conditional bias mask $O_{sp}^2$. Then, the second candidate mask is recalculated as:
\begin{equation}
\setlength{\abovedisplayskip}{7pt}
\setlength{\belowdisplayskip}{7pt}
    \hat{O}_{p}^{2}(\mathbf{p}) = O_{p}^2(\mathbf{p}) + b \cdot O_{sp}^2(\mathbf{p}).
\end{equation}
The final output can be obtained by the maximum argument between the first and second candidate masks.

\subsection{Extension to Class Incremental FSS}
Following the incremental learning in other few-shot tasks \cite{ganea2021incremental, perez2020incremental}, we give the definition of class incremental setting in few-shot segmentation. The Class Incremental Few-shot Semantic Segmentation (CIFSS) shares the same classes set with GFSS, in which there are base classes set $C_{\emph{base}}$ with adequate data available and novel classes set $C_{\emph{novel}}$ with a small amount of annotated data. Different from the one-off task of novel classes in GFSS, CIFSS focuses on the novel classes in a session stream. The model learns from the base session $S_{\emph{b}}$ with the base classes and adapts to novel classes with insufficient samples in sequential, \emph{i}.\emph{e}., the sequence session $\{S_{\emph{n}}^{1}, \cdots, S_{n}^{T}\}$ with the novel classes. It is critical to mention that we can only access to the data of \emph{t}-th session when attaching this session, and there is no overlap within these novel sessions. The performance of model with current session $S_{\emph{n}}^{\emph{t}}$ is evaluated on the test samples of $S_{\emph{n}}^{\emph{t}}$ and the previous sessions $\{S_{\emph{n}}^{1}, \cdots, S_{n}^{t-1}\}$. The GFSS can be viewed as a exception constructed with only base session and one novel session.

Owing to the extensible kernel and the mechanism of novel classes registration, the proposed model can be easy expanded to CIFSS. Similar to the training of GFSS, we learn the PKL and FCP module in the base session and all parameters are fixed during the sequence novel sessions. There are a series of novel classes registration processes corresponding to incremental classes learning with serialized novel sessions. To be specific, assume $K_{\emph{w}}^{t}$ is the whole-class kernels of \emph{t}-th session, the extended kernels of next session $K_{\emph{w}}^{t+1}$ is obtained by concatenating the novel session kernels $K_{\emph{n}}^{t+1}$:
\begin{equation}
\setlength{\abovedisplayskip}{7pt}
\setlength{\belowdisplayskip}{7pt}
    K_\emph{w}^{t+1} = \emph{Concat}([K_\emph{w}^{t}, K_\emph{n}^{t+1}]), \in \mathbb R^{(N_\emph{w}^{t}+N_\emph{n}^{t+1})\times C},
\end{equation}
where the calculation of $K_{\emph{n}}^{t+1}$ is the same with novel classes registration of GFSS referred to \ref{se.3.5.}. In the inference process, only the kernels of base session are passed to the PKL module, and all the kernels of few-shot sessions are treated as a novel whole in the CBBI.

\section{Experiments}
\label{sec:ex}
\subsection{Implementation Details}
\label{sec:ID}
{\bf Dataset.} Following \cite{tian2022generalized}, we evaluate the proposed method on two public datasets, \emph{i.e.}, PASCAL-$5^i$ and COCO-$20^i$. PASCAL-$5^i$ is modified from PASCAL VOC \cite{everingham2010pascal} with extra mask annotations from Semantic Boundaries Dataset (SBD) \cite{hariharan2011semantic} dataset, it has 20 object classes and one background class. COCO-$20^i$ consists of MS COCO \cite{lin2014microsoft} with 80 object classes and one background class for segmentation. In terms of GFSS, all the object classes are evenly divided into four folds: $\{5^i, i\in\{0,1,2,3\}\}$ for PASCAL-$5^i$ and $\{20^i, i\in\{0,1,2,3\}\}$ for COCO-$20^i$. The rest object classes as well as background are treated as base classes, where the object classes of specific fold are conducted as novel classes. Models are directly evaluated on the validation set with both base and novel classes simultaneously. In terms of CIFSS, it shares the same fold division with GFSS. The object classes of specific fold are evenly divided into five novel sessions, while the rest object classes and the background class are included into the base session. On PASCAL-$5^i$, there are 16 object classes in base session and 1 novel object class for each incremental session. As for COCO-$20^i$, the number of object classes in base session is 61 and novel object classes in each incremental session is 5. We evaluate the CIFSS models on the classes belonging to base session, current and previous incremental novel sessions with the validation set.

{\bf Evaluation metrics.} Similar to \cite{tian2022generalized}, we use mean Intersection over Union (\emph{m}IoU) for base classes as well as base session $\emph{m}\text{IoU}_{\mathcal{B}}$ and novel classes as well as incremental novel sessions $\emph{m}\text{IoU}_{\mathcal{N}}$. As for GFSS setting, we also provide the overall \emph{m}IoU with all classes denoted by $\emph{m}\text{IoU}_{\mathcal{O}}$. Since the base classes are the vast majority, the arithmetic mean might be dominated by $\emph{m}\text{IoU}_{\mathcal{B}}$ \cite{Baek_2021_ICCV}. We adopt the harmonic mean \emph{h}IoU of $\emph{m}\text{IoU}_{\mathcal{B}}$ and $\emph{m}\text{IoU}_{\mathcal{N}}$ which can emphasize the impact of novel classes.

{\bf Training details.} For fair comparison, the PSPNet \cite{PSP_CVPR_2017} with RestNet50 \cite{he2016deep} is adopted as our feature extractor, which is the same as most FSS and GFSS methods, and we train our model for 50 epochs both on PASCAL-$5^i$ and COCO-$20^i$. The model is optimized by the SGD with an initial learning rate of 2.5e-3, where momentum is 0.9 and the weight decay is set to 1e-4. All images together with the masks are all resized to 473$\times$473 for training and tested with their original sizes. The $\lambda$ of loss function is set to 0.6 and the bias value $b$ of CBBI is 0.5. We obtain the specific split results by averaging over five different random seeds, and the final reported results are averaged with all splits.

\subsection{Comparisons}

\begin{table}[]
\Large
\setlength{\abovecaptionskip}{0cm}
\setlength{\belowcaptionskip}{-0.3cm}
\caption{Quantitative results with GFSS setting on PASCAL-5$^i$ and COCO-20$^i$. Best-performing results are highlighted in bold.}
\begin{threeparttable}
\resizebox{1.04\linewidth}{!}{
\begin{tabular}{ccccccccc}
\toprule
\multirow{2}{*}{Methods} & \multicolumn{4}{c}{1-shot} & \multicolumn{4}{c}{5-shot} \\ 
\cmidrule(l){2-5} \cmidrule(l){6-9}
 & $\emph{m}\text{IoU}_{\mathcal{B}}$ & $\emph{m}\text{IoU}_{\mathcal{N}}$ & $\emph{m}\text{IoU}_{\mathcal{O}}$ & $\emph{h}\text{IoU}$ & $\emph{m}\text{IoU}_{\mathcal{B}}$ & $\emph{m}\text{IoU}_{\mathcal{N}}$ & $\emph{m}\text{IoU}_{\mathcal{O}}$ & $\emph{h}\text{IoU}$ \\ 
\midrule
\multicolumn{9}{c}{PASCAL-5$^i$} \\
\midrule
CANet \cite{CANet_CVPR_2019} & 8.73 & 2.42 & 7.23 & 3.79 & 9.05 & 1.52 & 7.26 & 2.60 \\
PFENet \cite{PFENet_TPAMI_2020} & 8.32 & 2.67 & 6.97 & 4.04 & 8.83 & 1.89 & 7.18 & 3.11 \\
SCL \cite{Zhang_CVPR_2021} & 8.88 & 2.44 & 7.35 & 3.83 & 9.11 & 1.83 & 7.38 & 3.05\\
PANet \cite{PANet_ICCV_2019} & 31.88 & 11.25 & 26.97 & 16.63& 32.95 & 15.25 & 28.74 & 20.85\\
BAM \cite{lang2022learning} & 64.65 & 17.06 & 53.32 & 27.00 & 65.28 & 19.99 & 54.50 & 30.61\\
CAPL \cite{tian2022generalized} & 65.48 & 18.85 & 54.38 & 29.27 & 66.14 & 22.41 & 55.72 & 31.96\\
Ours & \bf 68.84 & \bf 26.90 & \bf 58.86 & \bf 37.83 & \bf 69.22 & \bf 34.40 & \bf 61.18  & \bf 45.42 \\ 
\midrule
\multicolumn{9}{c}{COCO-20$^i$} \\
\midrule
Baseline \cite{tian2022generalized} & 36.68 & 5.84 & 29.06 & 10.01 & 36.91 & 7.26 & 29.59 & 12.13\\
BAM \cite{lang2022learning} & 44.08 & 6.74 & 34.86 & 11.69 & 46.10 & 11.21 & 37.49 & 18.03\\
CAPL \cite{tian2022generalized} & 44.61 & 7.05 & 35.46 & 12.18 & 45.24 & 11.05 & 36.80 & 17.76\\
Ours & \bf 46.36 & \bf 11.04 & \bf 37.71 & \bf 17.83 & \bf 46.77 & \bf 14.91 & \bf 38.90 & \bf 22.61\\
\bottomrule
\end{tabular}}
\end{threeparttable} 
\label{table_1}
\end{table}

\begin{table*}[th]
\setlength{\abovecaptionskip}{0cm}
\setlength{\belowcaptionskip}{-0.3cm}
\caption{1-shot Quantitative results with CIFSS setting on the PASCAL-5$^i$ and COCO-20$^i$. Best-performing results are highlighted in bold. $^*$ We use the weights of PSPNet \cite{PSP_CVPR_2017} backbone pretrained on base classes as the same as the BAM \cite{lang2022learning}.}
\resizebox{1.02\linewidth}{!}{
\begin{tabu}{c|c|c|ccc|ccc|ccc|ccc|ccc}
\tabucline[1pt]{-}
\multirow{2}{*}{Datasets} & \multirow{2}{*}{Methods} & sesson 0 & \multicolumn{3}{c|}{session 1} & \multicolumn{3}{c|}{session 2} & \multicolumn{3}{c|}{session 3} & \multicolumn{3}{c|}{session 4} & \multicolumn{3}{c}{session 5} \\ \cline{3-18} 
& & $\emph{m}\text{IoU}_{\mathcal{B}}$ & $\emph{m}\text{IoU}_{\mathcal{B}}$ & $\emph{m}\text{IoU}_{\mathcal{N}}$ & $\emph{h}\text{IoU}$ & $\emph{m}\text{IoU}_{\mathcal{B}}$ & $\emph{m}\text{IoU}_{\mathcal{N}}$ & $\emph{h}\text{IoU}$ & $\emph{m}\text{IoU}_{\mathcal{B}}$ & $\emph{m}\text{IoU}_{\mathcal{N}}$ & $\emph{h}\text{IoU}$ & $\emph{m}\text{IoU}_{\mathcal{B}}$ & $\emph{m}\text{IoU}_{\mathcal{N}}$ & $\emph{h}\text{IoU}$ & $\emph{m}\text{IoU}_{\mathcal{B}}$ & $\emph{m}\text{IoU}_{\mathcal{N}}$ & $\emph{h}\text{IoU}$ \\ \hline
\multirow{5}{*}{PASCAL-5$^i$} & PFENet$^*$ \cite{PFENet_TPAMI_2020} & 74.43 & 67.21 & 10.34 & 17.92 & 66.62 & 14.12 & 23.30 & 64.97 & 11.12 & 18.99 & 65.77 & 12.08 & 20.41 & 62.50 & 11.08 & 18.82\\
                              & iFS-RCNN \cite{nguyen2022ifs} & 72.43 & 67.71 & 11.08 & 19.04 & 67.17 & 16.63 & 26.66 & 65.51 & 12.94 & 21.61 & 65.48 & 11.75 & 19.92 & 63.36 & 12.39 & 20.73\\
                              & CAPL \cite{tian2022generalized} & 74.86 & 70.51 & 17.94 & 28.60 & 69.07 & 21.11 & 32.34 & 67.89 & 18.03 & 28.49 & 67.14 & 19.57 & 30.31 & 65.53 & 19.08 & 29.55 \\
                              & PIFS \cite{cermelli2021prototype} & 75.04 & 71.16 & 19.49 & 30.60 & 69.80 & 24.33 & 36.08 & 68.05 & 19.92 & 30.82 & 68.44 & 21.08 & 32.23 & 65.83 & 20.99 & 31.83\\
                             & Our& \bf 75.49 & \bf 72.93 & \bf 24.60 & \bf 36.79 & \bf 71.21 & \bf 29.32 & \bf 41.54 & \bf 70.76 & \bf 25.75 & \bf 37.76 & \bf 69.98 & \bf 26.11 & \bf 37.88& \bf 68.40 & \bf 27.28 & \bf 39.00 \\ \hline \hline
\multirow{5}{*}{COCO-20$^i$} & PFENet$^*$ \cite{PFENet_TPAMI_2020}& 54.11 & 45.79 & 4.33 & 7.91 & 44.58 & 5.18 & 9.28 & 43.07 & 3.36 & 6.23 & 42.40 & 5.64 & 9.96 & 41.99 & 5.05 & 9.02\\
                             & iFS-RCNN \cite{nguyen2022ifs} & 53.42 & 50.43 & 5.03 & 9.15 & 45.52 & 6.60 & 11.53 & 43.86 & 5.07 & 9.09 & 43.37 & 7.11 & 12.22 & 42.83 & 6.63 & 11.48\\
                             & CAPL \cite{tian2022generalized} & 54.43 & 50.25 & 6.69 & 11.81 & 46.67 & 7.04 & 12.23 & 45.92 & 5.53 & 9.87 & 44.08 & 7.77 & 13.21 & 43.85 & 7.51 & 12.82 \\
                             & BAM \cite{lang2022learning} & \bf 54.80 & \bf 52.04 & 8.83 & 15.10 & \bf 49.01 & 8.94 & 15.12 & 47.03 & 7.03 & 12.23 & 45.56 & 9.19 & 15.29 & 44.48 & 8.83 & 14.73\\
                             & PIFS \cite{cermelli2021prototype} & 54.27 & 51.59 & 8.90 & 15.18 & 48.52 & 9.53 & 15.93 & 46.96 & 6.94 & 12.09 & 45.69 & 9.93 & 16.31 & 44.27 & 8.80 & 14.68 \\
                             & Our & 54.39 & 51.78 & \bf 11.47 & \bf 18.55 & 48.33 & \bf 11.22 & \bf 18.08 & \bf 47.28 & \bf 9.61 & \bf 15.89 & \bf 46.25 & \bf 11.47 & \bf 18.33 & \bf 45.24 & \bf 10.74 & \bf 17.31 \\ \tabucline[1pt]{-}
\end{tabu}}
\label{table_2}
\end{table*}

{\bf GFSS Setting.} Table \ref{table_1} shows the 1-shot and 5-shot quantitative results of different methods on PASCAL-5$^i$ and COCO-20$^i$. It can be found that the proposed method shows substantial gains compared with the advanced approaches and achieves the state-of-the-art performance of all GFSS settings and datasets. As for the overall \emph{m}IoU on PASCAL-5$^i$, the proposed method exceeds the CAPL \cite{tian2022generalized} by 4.48\% and 5.46\% for 1-shot setting and 5-shot setting respectively. With the significant improvement on novel classes, our method surpasses the CAPL \cite{tian2022generalized} by 8.56\% and 13.46\% \emph{h}IoU, demonstrating the superiority of the proposed method. Meanwhile, better performance on novel classes can also weaken the confusion between base classes and novel classes and improve the $\emph{m}\text{IoU}_{\mathcal{B}}$.

{\bf CIFSS Setting.} We compare our method with the following modified approaches as: 1) the relevant incremental few-shot learning methods, such as PIFS \cite{cermelli2021prototype} and iFS-RCNN \cite{nguyen2022ifs}; 2) the current FSS and GFSS methods, \emph{i.e.}, PFENet \cite{PFENet_TPAMI_2020}, BAM \cite{lang2022learning} and CAPL \cite{tian2022generalized}. From Table \ref{table_2}, our method shows substantial gains compared with existing approaches. Since majority approaches adopt analogous backbone, we get similar performance in the session 0, which only construct with base classes. The superiority of our method both in base and novel classes continues to emerge as the novel session increased. It partly embodies the significant benefit of our old and novel classes knowledge ensemble mechanism, and partly verifies that the improvement on novel classes can result better performance of base classes as well.

\subsection{Ablation Studies}

\begin{table}[tp]
\Large
  \centering  
%   \fontsize{7}{7}\selectfont
  %\small
  \setlength{\abovecaptionskip}{0cm}
  \setlength{\belowcaptionskip}{-0.3cm}
  \caption{Ablation Study of the effect with different components.}
  \begin{threeparttable}
  \resizebox{1.02\linewidth}{!}{
  \begin{tabular}{ccccccccc}
  \toprule
  \multirow{2}{*}{PKL} & \multirow{2}{*}{FCP} & \multirow{2}{*}{CBBI} & \multicolumn{3}{c}{1-shot} & \multicolumn{3}{c}{5-shot} \\ 
  \cmidrule(l){4-6} \cmidrule(l){7-9}
  \multicolumn{3}{c}{} & $\emph{m}\text{IoU}_{\mathcal{B}}$ &$\emph{m}\text{IoU}_{\mathcal{N}}$& $\emph{h}\text{IoU}$ &$\emph{m}\text{IoU}_{\mathcal{B}}$ & $\emph{m}\text{IoU}_{\mathcal{N}}$ & $\emph{h}\text{IoU}$\cr
  \midrule
  & & & 61.22 & 14.50 & 23.45 & 62.07 & 19.73 & 29.94 \cr
  \checkmark & & & 66.82 & 22.95 & 34.17 & 67.73 & 29.18 & 40.79 \cr
  & \checkmark & & 64.49 & 16.73 & 26.57 & 65.28 & 23.39 & 34.44 \cr
  \checkmark & \checkmark &  & 67.73 & 23.58 & 34.98 & 68.05 & 31.45 & 43.02 \cr
  \checkmark & \checkmark & \checkmark & \bf 68.84 & \bf 26.90 & \bf 37.83 & \bf 69.22 & \bf 34.40 & \bf 45.42 \cr
  \bottomrule
  \end{tabular}}
  \end{threeparttable} 
   \label{table_3}
\end{table}

\begin{figure*}[ht!]
    \centering
    \setlength{\abovecaptionskip}{0cm}
	\includegraphics[width=\textwidth]{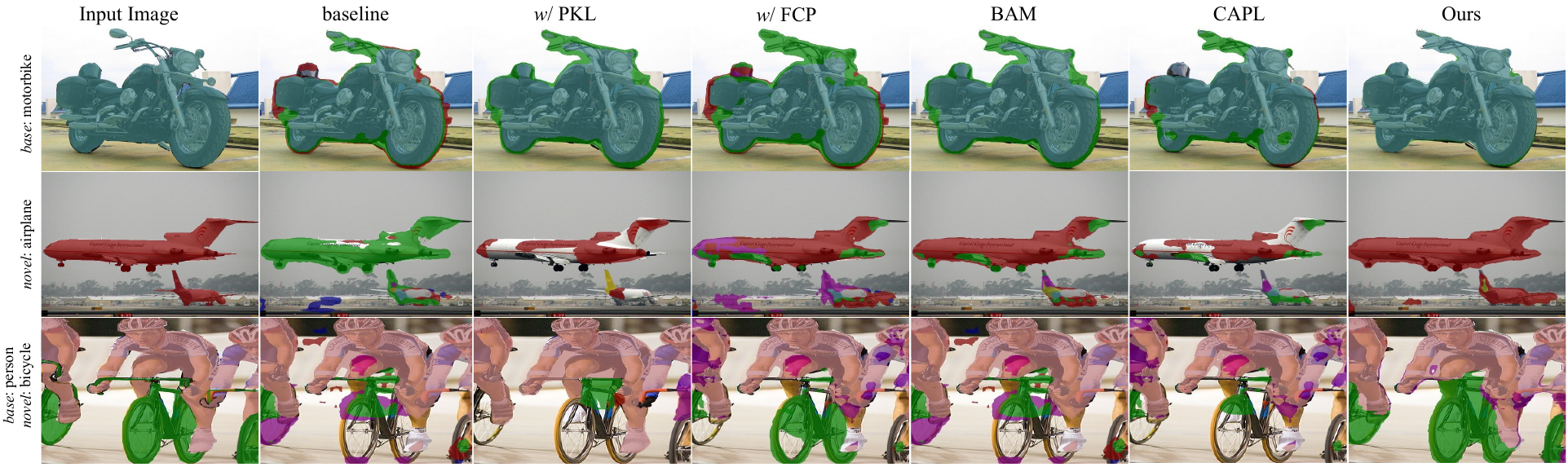}
	\caption{Qualitative results of our method, ablation of each components, BAM \cite{lang2022learning} and CAPL \cite{tian2022generalized} on PASCAL-5$^i$ benchmark.}
 % Our approach can assign correct labels to both the base classes and novel classes.
 	\label{fig5_qualitative_results}
\end{figure*}

\noindent \textbf{Component-wise Effectiveness.} We ablate the three major components of proposed method in Table \ref{table_3}: prototypical kernel learning module, foreground contextual perception module and conditional bias based inference. We use the PSPNet \cite{PSP_CVPR_2017} modified with learnable kernels as the baseline, in which simple concat operation is conducted between base-class and novel-class kernels. We can see that adding either one of PKL module or FCP module brings substantial improvement compared with the baseline. It means the proposed two modules are indeed beneficial to the novel classes learning in few-shot scenarios, and can mitigate the knowledge division with the extremely class-imbalanced GFSS setting. Moreover, it worth noting that the FCP module explicit plays a vital role in providing the general foreground information, the implicit effect is to boost the feature expression of unknown targets. Fig. \ref{fig5_qualitative_results} shows the qualitative results, which straightforward embodies the effectiveness of proposed method.

\noindent \textbf{Prototypical Kernel Learning Module.} The motivation of PKL module is to empower the prototypical ability for base-class kernels, so as to maintain the consistency with the registered novel classes, which is also prototypical aggregated from limited data. We replace the PKL with directly adopting the base-class kernels to the input feature maps as our baseline. As shown in the upside of Table \ref{table_4}, we first construct the ablation studies on the adaptation learning rate, it is clear that the fixed learning rate brings worse results with the distraction of noisy gradient signals during the prototypical kernel update. Meanwhile, we introduce several alternative prototype fusion approaches which are performed by the adaptive pooling (AP) or liner projection (LP) after the channel-wise \emph{concat} operation between $K_b$ and $P_{\emph{mb}}$. The results show that our method with the proposed prototypical kernel update remarkably outperforms the aforementioned approaches. It can strike a better balance between generality and specificity, especially for novel targets in generalized inference process, thus is more suitable to the GFSS.

\begin{table}[]
\Large
\setlength{\abovecaptionskip}{0cm}
\setlength{\belowcaptionskip}{-0.2cm}
  \caption{Ablation performances of the proposed method. \emph{fixed}ALR means using fixed
learning rate referred to Eq. (\ref{Eq.3}). $\text{pixel}_{\emph{AP}\rightarrow \emph{MP}}$ indicates replacing average pooling with max pooling in pixel dimension referred to Eq. (\ref{Eq.6}), and $\text{image}_{\emph{MP}\rightarrow \emph{AP}}$ is the similar replacement in image dimension. \emph{u}CBBI and \emph{c}CBBI denote the unconditional form and learnable convolution form for CBBI.}
\resizebox{1.02\linewidth}{!}{
\begin{tabular}{ccccccc}
\toprule
\multirow{2}{*}{Settings} & \multicolumn{3}{c}{1-shot} & \multicolumn{3}{c}{5-shot} \\ 
\cmidrule(l){2-4} \cmidrule(l){5-7}
 & $\emph{m}\text{IoU}_{\mathcal{B}}$ & $\emph{m}\text{IoU}_{\mathcal{N}}$ & $\emph{h}\text{IoU}$ & $\emph{m}\text{IoU}_{\mathcal{B}}$ & $\emph{m}\text{IoU}_{\mathcal{N}}$ & $\emph{h}\text{IoU}$ \\ 
\midrule
\emph{w/o} PKL & 64.49 & 16.73 & 26.57 & 65.28 & 23.39 & 34.44\\
\emph{fixed}ALR & 66.90 & 20.48 & 31.36 & 67.72 & 26.08 & 37.66\\
concat+AP & 65.91 & 19.41 & 30.00 & 66.86 & 25.51 & 36.93\\
concat+LP & 66.27 & 21.19 & 32.11 & 67.71 & 26.89 & 38.49\\
\midrule
\midrule
\emph{w/o} FCP & 66.82 & 22.95 & 34.17 & 67.73 & 29.18 & 40.79\\
$\text{pixel}_{\emph{AP}\rightarrow \emph{MP}}$ & 65.94 & 22.31 & 33.34 & 66.42 & 28.91 & 40.28\\
$\text{image}_{\emph{MP}\rightarrow \emph{AP}}$ & 65.05 & 21.94 & 32.81 & 66.18 & 28.30 & 39.65\\
GradCAM \cite{selvaraju2017grad} & 66.94 & 23.36 & 34.74 & 68.08 & 29.80 & 41.46\\
% ScoreCAM & 67.11 & 23.49 & 36.28 & 68.21 & 30.12 & 41.79\\
PAGENet \cite{wang2019salient} & 67.57 & 24.80 & 36.28  & 68.83 & 31.70 & 43.41\\
MINet \cite{pang2020multi} & 67.89 & 24.99 & 36.53 & 68.65 & 31.47 & 43.16\\
\midrule
\midrule
\emph{w/o} CBBI & 67.73 & 23.58 & 34.98 & 68.05 & 31.45 & 43.02\\
\emph{u}CBBI & 66.29 & 20.83 & 31.70 & 67.13 & 25.38 & 36.83\\
\emph{c}CBBI & 65.53 & 18.86 & 29.29 & 66.90 & 23.39 & 34.66\\
Ours & 68.84 & 26.90 & 37.83 & 69.22 & 34.40  & 45.42 \\
\bottomrule
\end{tabular}
}
\label{table_4}
\end{table}

\noindent \textbf{Foreground Contextual Perception module.} The FCP aims to offer a general foreground detection which helps to the mining of novel targets. Similarly, the model without FCP is conducted as the baseline, as well as replacing the CBBI with normal inference. The middle part of Table \ref{table_4} demonstrates the ablation results of the pixel-wise group correlation aggregation referred to Eq. (\ref{Eq.6}), which is the key of FCP, as well as the comparison with other alternative foreground detection approaches. Using average pooling on pixel dimension of the pixel-wise group correlation tends to introduce more background noisy, leading to worse performance. We also compare our FCP with CAM-based and well-designed salient object detection approaches. Since the CAM-based methods are heavily relied on the extra classes information, such as class-wise gradients \cite{selvaraju2017grad} or activation values \cite{wang2020score}, it lacks the ability of novel classes detection. Similar deficiency also lies in other well-designed salient detection methods, \emph{e.g.}, PAGENet \cite{wang2019salient} and MINet \cite{pang2020multi}, which only consider the interaction within specific classes and lose the generality of foreground detection. The FCP pays more attention to the robust and universal feature patterns with abundant pseudo episode, thus is more easy to generalize to novel targets.

\noindent \textbf{Conditional Bias Based Inference.} In our method, the CBBI plays a role in combining the outputs of the PKL module and the FCP module. We also experiment on the way of fusing those outputs in the bottom part of Table \ref{table_4}. When the class-agnostic foreground mask predicted by the FCP module is utilized with unconditional form, the performance gets worse with the interference of background prediction. We also replace the strategy based inference with learnable convolution structure with end-to-end training, the underperform results indicate that introducing parameterized inference module may heavy the optimization of model and even block the learning of other modules.

\noindent \textbf{Performance distinction between PASCAL-5$^i$ and COCO-20$^i$.} We further investigate the improvement gap between PASCAL-5i and COCO-20i referred to Table \ref{table_1}, \emph{e.g.}, 8.05\% v.s. 3.99\% exceed CAPL \cite{tian2022generalized} on mIoU of novel classes. The most obvious difference between these two datasets is that coco-20i contains larger classes set than PASCAL-5$^i$. Thus, we design the ablation experiments with gradually enlarging the subset of COCO, which is started from a similar scale with PASCAL-5$^i$. As shown in Table \ref{table_5}, we find parallel improvements are raised when the scale and distribution of classes are similar to PASCAL-5$^i$, and as the number of classes increasing, the gaps are reduced. Comparable phenomenon also happens on the contrast between the CAPL \cite{tian2022generalized} and baseline, indicating the common issue that the effect of GFSS model may get weakened as the scale of classes set increases.

\begin{table}[]
% \Large
\setlength{\abovecaptionskip}{0.1cm}
\setlength{\belowcaptionskip}{-0.3cm}
\caption{Results on the subset of COCO. \emph{a}+\emph{b}/\emph{c} indicates the combination of classes, where \emph{a} is the number of base classes, \emph{b}=1 represents the background class and \emph{c} is the number of novel classes.}
\begin{threeparttable}
\resizebox{1.02\linewidth}{!}{
\begin{tabular}{ccccccccc}
\toprule
\multirow{2}{*}{Methods} & \multicolumn{2}{c}{15+1/5} & \multicolumn{2}{c}{30+1/10} &  \multicolumn{2}{c}{45+1/15} & \multicolumn{2}{c}{60+1/20}\\ 
\cmidrule(l){2-3} \cmidrule(l){4-5} \cmidrule(l){6-7} \cmidrule(l){8-9}
 & $\emph{m}\text{IoU}_{\mathcal{B}}$ & $\emph{m}\text{IoU}_{\mathcal{N}}$ & $\emph{m}\text{IoU}_{\mathcal{B}}$ & $\emph{m}\text{IoU}_{\mathcal{N}}$ & $\emph{m}\text{IoU}_{\mathcal{B}}$ & $\emph{m}\text{IoU}_{\mathcal{N}}$ & $\emph{m}\text{IoU}_{\mathcal{B}}$ & $\emph{m}\text{IoU}_{\mathcal{N}}$\\ 
\midrule
Baseline \cite{tian2022generalized} & 51.19 & 10.08 & 47.74 & 8.22 & 41.37 & 7.10 & 36.68 & 5.84\\
CAPL \cite{tian2022generalized} & 63.39 & 16.08 & 57.22 & 13.39 & 50.08 & 10.16 & 44.61 & 7.05\\
Ours & 66.70 & 23.76 & 59.48 & 18.91 & 52.01 & 14.27 & 46.36 & 11.04\\
\midrule
\rowcolor{lightgray!40} Improvement & +3.31 & +7.76 & +2.26 & +5.52 & +1.92 & +4.11 & +1.75 & +3.99\\
\bottomrule
\end{tabular}}
\end{threeparttable} 
\label{table_5}
\end{table}

\section{Conclusion}
In this work, we have proposed to leverage the prototypical kernel learning and open-set foreground perception for generalized few-shot semantic segmentation. Special to two of major issues existed in current GFSS, \emph{i.e.}, representation division and embedding prejudice. We offer an effective scheme with three components, which are PKL, FCP and CBBI. To maintain the prototypical representation consistency and prevent novel targets from being misclassified as background, the PKL and FCP are proposed to perform adaptive prototypical kernel update and open-set targets detection, respectively. Moreover, the CBBI is designed for the ensemble of the aforementioned two modules and offers the final prediction. Besides, we further extend the GFSS to CIFSS which is more challenging with incremental stream of novel classes. Extensive experiments demonstrate that the proposed method achieves state-of-the-art results. 

{\small
\bibliographystyle{ieee_fullname}
\bibliography{egbib}
}

\end{document}